\documentclass[letterpaper, 10 pt, journal, twoside]{ieeetran}

\IEEEoverridecommandlockouts                              

\usepackage{times}
\usepackage{epsfig}
\usepackage{graphicx}
\usepackage{amsmath}
\usepackage{amssymb}
\usepackage{booktabs}
\usepackage{multirow}
\usepackage{float}
\usepackage{mathtools}
\usepackage{adjustbox}
\usepackage{multicol}
\usepackage{wrapfig}
\usepackage{sidecap}
\usepackage{hyperref}

\hypersetup{
    colorlinks=true,
    urlcolor=blue
}

\title{
Don't Forget The Past: Recurrent Depth Estimation from Monocular Video
}

\author{Vaishakh Patil$^{1}$, Wouter Van Gansbeke$^{2}$, Dengxin Dai$^{1}$ and Luc Van Gool$^{1,2}$

\thanks{Manuscript received: February, 24, 2020; Revised May, 30, 2020; Accepted July, 20, 2020.}
\thanks{This paper was recommended for publication by Editor Cesar Cadena upon evaluation of the Associate Editor and Reviewers' comments.}
\thanks{This work is supported by Toyota Motor Europe.}
\thanks{$^{1}$Vaishakh Patil, Dengxin Dai and Luc Van Gool are with the Toyota TRACE-Zurich at the Computer Vision Lab, ETH Zurich, 8092 Zurich, Switzerland {\tt\footnotesize $\{$patil, dai, vangool$\}$@vision.ee.ethz.ch}}%
\thanks{$^{2}$ Wouter Van Gansbeke and Luc Van Gool are with the Toyota TRACE-Leuven at the
Dept. of Electrical Engineering ESAT, KU Leuven, 3001 Leuven, Belgium
        {\tt\footnotesize  $\{$wouter.vangansbeke, luc.vangool$\}$@kuleuven.be}}%
\thanks{Digital Object Identifier (DOI): see top of this page.}
}

\begin{document}

\maketitle

\begin{abstract}
Autonomous cars need continuously updated depth information. Thus far, depth is mostly estimated independently for a single frame at a time, even if the method starts from video input. Our method produces a time series of depth maps, which makes it an ideal candidate for online learning approaches. 
In particular, we put three different types of depth estimation (supervised depth prediction, self-supervised depth prediction, and self-supervised depth completion) into a common framework. We integrate the corresponding networks with a ConvLSTM such that the spatiotemporal structures of depth across frames can be exploited to yield a more accurate depth estimation. Our method is flexible. It can be applied to monocular videos only or be combined with different types of sparse depth patterns. We carefully study the architecture of the recurrent network and its training strategy. We are first to successfully exploit recurrent networks for real-time self-supervised monocular depth estimation and completion. Extensive experiments show that our recurrent method outperforms its image-based counterpart consistently and significantly in both self-supervised scenarios. It also outperforms previous depth estimation methods of the three popular groups. Please refer to \href{https://www.trace.ethz.ch/publications/2020/rec\_depth\_estimation/index.html}{our webpage} for details.
\end{abstract}



\section{Introduction}
\label{sec:intro}

\IEEEPARstart{H}{igh} precision depth estimation is essential for a variety of applications such as augmented reality, autonomous vehicles, and mobile robots. The last years have witnessed tremendous progress in depth estimation, especially after the wide deployment of deep neural networks.
On one hand, supervised learning algorithms are constantly improving for depth estimation from RGB images~\cite{kitti,Eigen:2014,deeper:depth:3dv16,dorn:cvpr18}. On the other hand, self-supervised depth estimation from camera motion are also steadily improving~\cite{unsupervised:depth:ego:cvpr17,ddvo:cvpr18,geonet:cvpr18,dfnet:eccv18}. Recently, several studies on depth completion were made, aiming at completing the depth map obtained by a high-end LiDAR sensor, namely HDL-64E, by using a paired image for guidance~\cite{uhrig:sparsity:3dv17,van:sparse:mva19,ma:self:sparse:icra19,jaritz:sparse:3dv18,cheng:depth:sparse:eccv18, eldesokey:confidence:sparse:18}.

Whereas great progress is being made in all three cases, none of those methods would seem optimal for mobile robot applications though. As they move,  mobile platforms - be it cars or assistive robots - perceive the world more as a video stream than as isolated images. While videos are used in the training stage of self-supervised depth prediction methods for computing the view-synthesis loss across neighboring frames~\cite{unsupervised:depth:ego:cvpr17,ddvo:cvpr18,geonet:cvpr18,monodepth2:iccv19}, they ignore the intrinsic temporal dependency across frames at testing. The perceptual inputs along a trajectory and the underlying scene geometries are highly correlated~\cite{see:by:moving:iccv15}. Given our context of robotic applications, we propose a depth recovery method that both trains and tests on time series of data. This way, the perception-geometry correlation can be leveraged the best.

\begin{figure}[!t]
\begin{center}
 \includegraphics[width=0.8\linewidth]{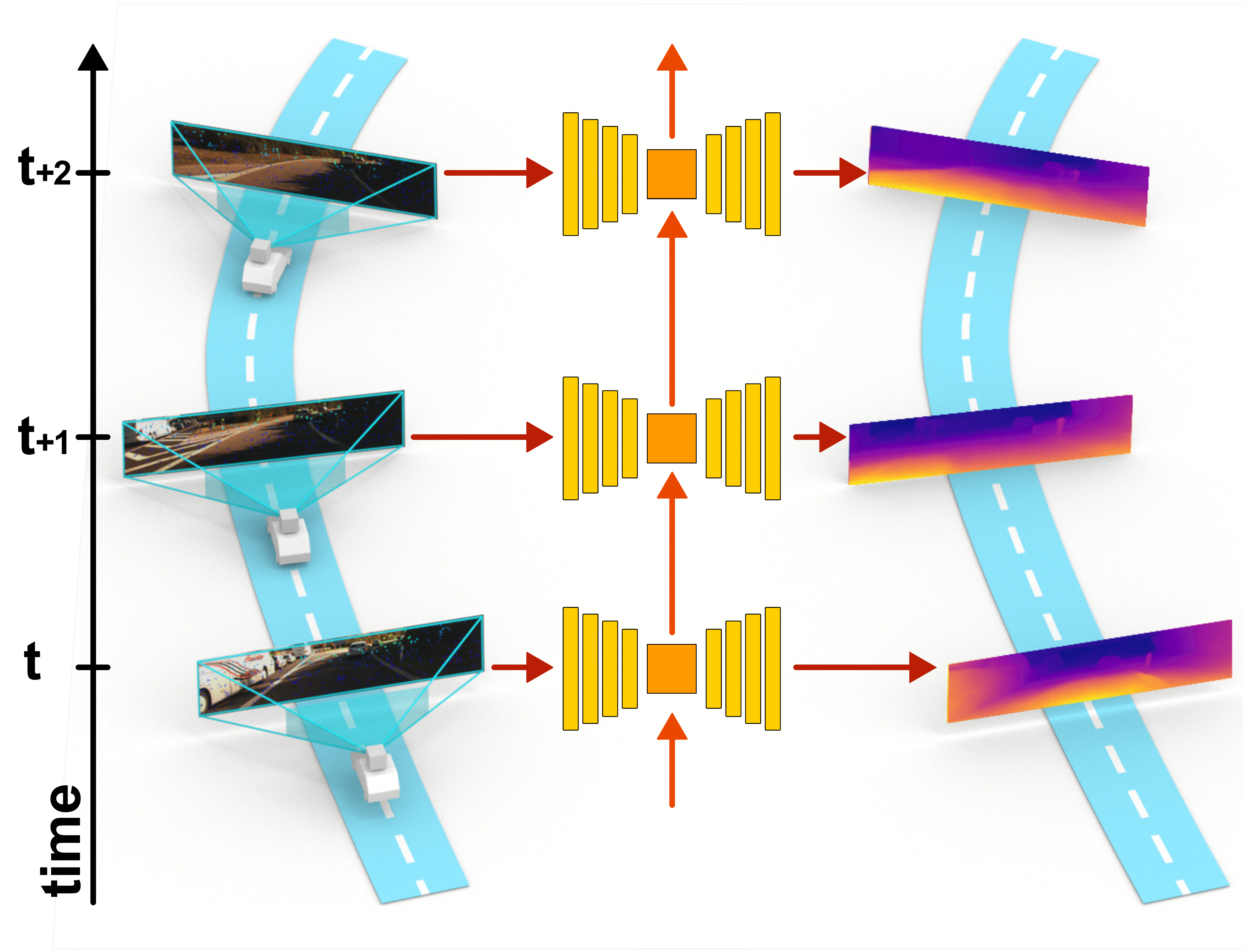}
\end{center}
\vspace{-2mm}
 \caption{Our method trains and tests on time series of data and produces accurate depth maps for robotic platforms which perceive the world more as a video stream than as isolated images.}
\label{fig:intro}
 \vspace*{-5mm}
\end{figure}

Also, none of the three settings seem to be optimal by themselves. The setting of depth prediction from RGB images is cheap but requires large training datasets with accurate ground truth; the setting for self-supervised depth estimation from videos leverages ego-motion information (to some extent) but have yet to generate the best results; and the setting for depth completion with a LiDAR sensor and a camera yields good results but is quite costly.

Hence, in this work, we put the three different types of depth estimation, i.e. supervised depth prediction from RGB images~\cite{deeper:depth:3dv16,dorn:cvpr18}, self-supervised depth prediction with monocular videos~\cite{geonet:cvpr18,monodepth2:iccv19} and self-supervised depth completion~\cite{ma:self:sparse:icra19} but with monocular videos, into a common framework, and then integrate their corresponding `backbone' networks with a convolutional LSTM (ConvLSTM) such that spatiotemporal information across frames can be exploited for more accurate depth estimation.


ConvLSTMs have been designed to exploit temporal information but it is still unclear how they can be trained properly for self-supervised depth estimation from current literature. We claim to be the first to propose an novel and effective strategy to integrate ConvLSTM into the unified depth estimation pipeline. The training is challenging because 1) the size of the feature maps is large for dense prediction which limits the sequence length due to memory issues; and 2) under the standard training strategy, ConvLSTM based networks need a long sequence to learn the hidden state properly. Our training strategy addresses these issues.



In summary, this work makes three contributions: 1) a novel recurrent network to exploit spatiotemporal information for depth estimation,
2) an innovation to effectively train a ConvLSTM based network for dense prediction tasks with video inputs; and
3) extensive experiments and detailed ablation studies. Experiments show that our recurrent method outperforms its image-based counterpart and the current SOTA methods consistently and significantly in all the considered scenarios.

\section{Related Work}
\label{sec:related}

\noindent

\textbf{Supervised Depth Estimation from RGB Images}.
A large body of work focuses on depth estimation from images with varying settings: from using image pairs~\cite{depth:geometry:rescue:eccv16,unsupervised:depth:left:right:cvpr17}, to using multiple overlapping images captured from different viewpoints~\cite{multi:view:stereo:15}
Here we summarize work related to the supervised learning of depth from a single RGB image. \cite{make3d} is among the earliest work popularizing this idea. Local image statistics are used to infer 3D planes for local patches and the final results are optimized globally over a defined Markov Random Field. Later on, deep convolutional neural networks were used for this task~\cite{Eigen:2014,Eigen:2015,depthfield:single:image:pami16,deeper:depth:3dv16}. The research focus was mainly on improving the network architecture~\cite{Eigen:2014,deeper:depth:3dv16}, formulating multi-task learning~\cite{Eigen:2015}, and combining CNNs and CRFs~\cite{depthfield:single:image:pami16}.
In order to alleviate the dependency on large-scale ground-truth depth images, methods that learn directly from stereo pairs were developed~\cite{unsupervised:depth:left:right:cvpr17}. The core idea is to use  left-right view similarity as the supervisory signal. This line of work has been further extended to a semi-supervised setting~\cite{semi:depth:cvpr17}, where direct supervision from LiDAR sensors and indirect supervision from image warping are combined.

\textbf{Self-supervised Depth Estimation from Videos}.
To lower the dependency on ground truth depth images, many recent works have shifted the focus to self-supervised depth estimation from monocular videos by using view-synthesis or its variants as the supervisory signal~\cite{unsupervised:depth:ego:cvpr17,ddvo:cvpr18,geonet:cvpr18,depth:odometry:cvpr18,dfnet:eccv18,monodepth2:iccv19, gordon:iccv2019}.
While promising results have been shown, the training of self-supervised methods requires careful hyper-parameter tuning and suffers from scale ambiguity, which needs to be addressed, e.g. by using stereo images~\cite{depth:odometry:cvpr18} or by data normalization~\cite{ddvo:cvpr18}. While consecutive video frames are used for the view-synthesis loss, the spatiotemporal information is not exploited, especially at testing time.


\textbf{Depth Completion}.
While steady progress has been made for depth estimation from RGB images, the performance can be improved when assisted by other sources. One notable example is that of sparse depth inputs, either from cheap LiDAR sensors~\cite{geometry:line:icra17} or from SLAM or structure-from-motion systems~\cite{sparse:to:dense:icra18,depth:rgb:sparse:eccv18}.
There has been a large body of work~\cite{uhrig:sparsity:3dv17,van:sparse:mva19,ma:self:sparse:icra19, jaritz:sparse:3dv18,cheng:depth:sparse:eccv18, eldesokey:confidence:sparse:18} developed for the task of depth completion defined by the KITTI Depth Completion Benchmark~\cite{uhrig:sparsity:3dv17}. Methods have also been developed for depth completion with sparse Radar points recently \cite{radar:depth:20}. The main research focus of this strand is how to spatially propagate Automotive LiDAR depth data under image guidance. The established knowledge, such as the design of network architectures for spatial propagation, can be borrowed to design our recurrent method. The main focus of our work, however, lies in how to fuse or propagate depth information over frames.


\textbf{Depth Estimation for Videos}.
Our method is designed for online depth estimation in videos. Similar idea of online estimation is proposed in recent works \cite{zhang:iccv19:exploiting, lstm:cvprw18}, where they use ConvLSTM but in supervised framework. There are also earlier methods for offline depth estimation from videos~\cite{depth:transfer:video:14}, in which local motion cues and optical flow are used to produce temporally consistent depth maps.



\section{Approach}
\label{sec:approach}
As stated in Sec.~\ref{sec:intro}, depth estimation has been tackled under multiple settings, each has its own strengths and weaknesses. These systems, however, are mostly image-based and lack the capability of integrating information over video sequences obtained by moving robotic platforms. In this section, we first summarize three existing depth estimation methods in a unified formulation and then define our recurrent framework for learning time series of depth maps for all the three methods.


Before going into the details, we define certain notations used by the methods. Let us denote by $I(x,y)$ the RGB vector-valued image, $\bar{D}(x,y)$ the input sparse depth, and $D(x,y)$ the ground truth depth map; all three have the same dimensions $H \times W$. While $I(x,y)$ is dense, $D(x,y)$ has regions with missing values which are indicated by zero. $\bar{D}(x,y)$ is much sparser compared to $D(x,y)$; it is usually a sub-sampled depth map from $D(x,y)$ to simulate sparse input patterns that can be obtained via a low resolution LiDAR sensor. The detailed sampling procedures for varying sparse input patterns are given in Sec.~\ref{sec:self:depth:comp}.
The timestamp of time series data is denoted  by $t$ in subscript, where $t \in \{1,2, ..., T\}$ with $T$ the total number of frames of a data sequence. We assume that the image and the depth maps are synchronized throughout the sequence.

\subsection{Supervised Depth Prediction with RGB Images}
\label{sec:single:image}
Supervised depth prediction with monocular RGB images has been a popular research topic. Tremendous progress has been made in the past years~\cite{dorn:cvpr18, yang:eccv18, Eigen:2015, Eigen:2014, guo:eccv18}.
The task is to learn a function $f: I \rightarrow \hat{D}$, where $\hat{D}$ has the same resolution $H \times W$ and has predictions for all pixel coordinates $(x,y)$ where $x \in \{1, .., H\}$ and $y \in \{1, ..., W\}$. We represent the depth estimation network as an encoder-decoder type of network architecture. Then, the learning takes the form:
\vspace{-11mm}
\begin{multicols}{2}
    \begin{equation}
    \label{eq:encoder}
        X=f_{\text{encoder}}(I),
    \end{equation} \break
    \begin{equation}
    \label{eq:decoder}
        \hat{D}=f_{\text{decoder}}(X),
    \end{equation}
\end{multicols}
\vspace{-2mm}
where $X$ is the summarized representation by the encoder, which is compact and will be used to pass information across video frames for depth estimation from a video sequence as presented in Sec.~\ref{sec:monocular:video}.

We follow~\cite{deeper:depth:3dv16} and use the berHu loss, which gives slightly better results than the L1 and L2 loss. Let us define a binary mask $M(x,y)$ of dimensions $H \times W$, where $M(x,y)=1$ defines $(x,y)$ locations of valid values for the ground truth depth map $D(x,y)$.  The loss can then be formulated as
\vspace{-2mm}
\begin{equation}
    \label{eq:supervised:objective}
    \mathcal{L}_{\text{berHu}} =   \sum_{t} \|M_t \circ D_t - M_t \circ \hat{D}_t \|_{\delta},
\end{equation}
where `$\circ$' denotes the Hadamard product in order to ignore the invalid pixels of the ground truth depth picture $D_t$ when computing the loss, and $\|.\|_\delta$ is set by following~\cite{deeper:depth:3dv16}.


\subsection{Self-supervised Depth Prediction with Monocular Video}
\label{sec:self:depth:pred}
Self-supervised depth estimation from monocular video has been quite successful in recent years~\cite{monodepth2:iccv19, depth:geometry:rescue:eccv16, unsupervised:depth:left:right:cvpr17, unsupervised:depth:ego:cvpr17}. We will mostly follow the presentation of the state-of-the-art method Monodepth2~\cite{monodepth2:iccv19} in this section. We represent the depth estimation network with the same encoder-decoder network as defined in Eq.~\ref{eq:encoder} and Eq.~\ref{eq:decoder}. Since there is no ground-truth depth map $D(.)$, the view-synthesis loss is used instead of the standard supervised loss functions.

In particular, if $K$ denotes the camera intrinsic matrix, and $\Phi_{t \rightarrow t+\Delta t}$  the relative camera pose from view $t$ to a neighboring view $t+\Delta t$, the warped image is:
\vspace{-2mm}
\begin{equation}
\label{eq:warping2}
    I_{t + \Delta t \rightarrow t} = {I}_{t+\Delta t}\left<  K\Phi_{t \rightarrow t+\Delta t}\hat{D}_tK^{-1}\right>,
\end{equation}
where $\left<.\right>$ is a bilinear sampling function to sample the source image. The view-synthesis loss of our method is defined as
\vspace{-2mm}
\begin{equation}
\label{eq:vsloss:frame}
    \begin{split}
        \mathcal{L}_{{vs(t +\Delta t \rightarrow t)}} =  \frac{\alpha}{2}(1-SSIM(I_{t}, I_{t+\Delta t \rightarrow t})) \\
        + (1 - \alpha)\|I_{t} - I_{t+\Delta t \rightarrow t}\|_1,
    \end{split}
\end{equation}
where $\alpha= 0.85$.

\begin{figure}[!t]
\centering
\includegraphics[width=1.0\linewidth, angle=0]{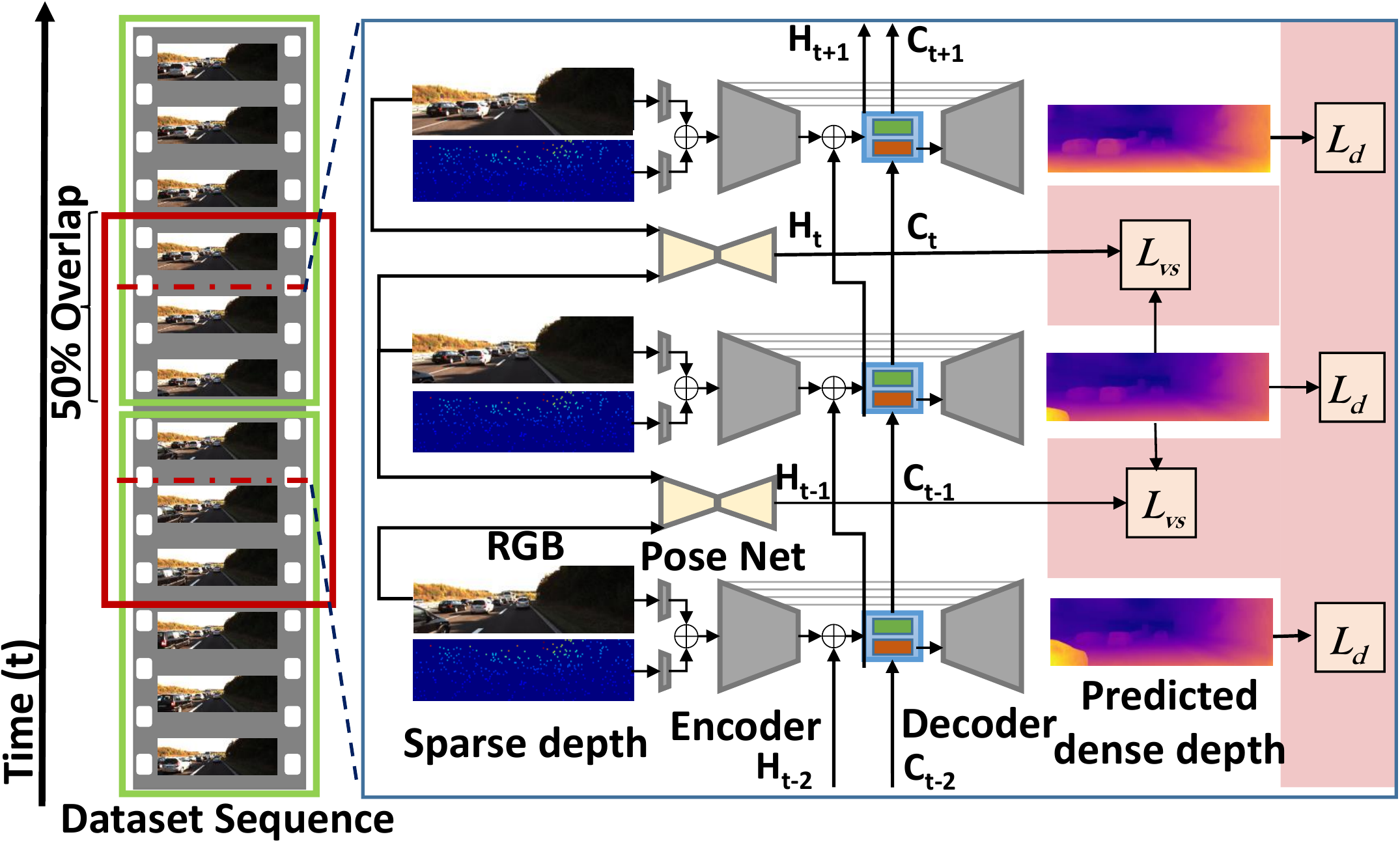}
\caption{The pipeline of our recurrent learning framework for depth recovery with monocular video and sparse depth sensing.}
\label{fig:pipeline}
 \vspace*{-5mm}
\end{figure}

In addition to estimating depth, the model also needs to estimate the camera pose $\Phi_{t \rightarrow t+\Delta t}$. This involves training a pose estimation network that predicts the corresponding camera transformations with the same sequence of frames as input.
The $\mathcal{L}_{{vs(t +\Delta t \rightarrow t)}}$ is computed at multiple scales of the decoder, similar to~\cite{monodepth2:iccv19}.

The final view-synthesis loss is aggregated over all considered source (neighboring) frames. In this work, $\Delta t \in \{-1, 1\}$, i.e. for each frame $t$, the previous frame $t-1$ and the next frame $t+1$ are used to compute the loss.
Following~\cite{monodepth2:iccv19}, at each pixel, we use the minimum photometric error over all source images. Thus, the final view-synthesis loss is
\vspace{-2mm}
\begin{equation}
\label{eq:vsloss:final}
        \mathcal{L}_{vs} = \min_{\Delta t \in \{-1, 1\}} \mathcal{L}_{{vs(t+\Delta t \rightarrow t)}}.
\end{equation}

Following~\cite{monodepth2:iccv19}, we also use the edge-aware smoothing loss:
\vspace{-3mm}
\begin{equation}
    \label{eq:smooth}
    \mathcal{L}_{smooth} = |\partial_x \hat{D}^*_t| e^{ \|-\partial_x I_t\|} + |\partial_y \hat{D}^*_t| e^{ \|-\partial_y I_t\|},
\end{equation}
where $\hat{D}^*_t$ is mean-normalized $\hat{D}_t$ to avoid shrinking the depth values. Our learning algorithm is trained with a combined loss:
\vspace{-3mm}
\begin{equation}
    \mathcal{L}_{\text{self\_pred}} = \mu\mathcal{L}_{vs} +  \nu\mathcal{L}_{smooth}.
\end{equation}
$\mu$ represents the pixel-wise masking of the view-synthesis loss to ignore certain objects. This addresses the problem that a car, travelling at the same speed as the camera, will be predicted at infinite depth \cite{monodepth2:iccv19}.

\subsection{Self-supervised Depth Completion with Monocular Video and Sparse Depth Maps}
\label{sec:self:depth:comp}
Supervised depth prediction methods (Sec.~\ref{sec:single:image}) are able to achieve good performance but require large training sets with accurate ground truth depth and have difficulty to generalize to new scenarios. Self-supervised depth prediction methods (Sec.\ref{sec:self:depth:pred}) are easy to `generalize', but have yet to yield the state-of-the-art results. In this section, we present another stream of method called self-supervised depth completion following~\cite{ma:self:sparse:icra19}.

The method requires a monocular video $(I_1, I_2, ..., I_T)$ and synchronized sparse depth maps $(\bar{D}_1, \bar{D}_2, ..., \bar{D}_T)$ as inputs. While it is hard to obtain dense depth map $D_t$, sparse depth map $\bar{D}_t$ are relatively cheap and easy to acquire, e.g.~via 2D LiDAR sensors. Compared to self-supervised depth prediction, this vein of research also focuses on developing a suitable network architecture to better fuse the information from these two modalities. Typical examples include a simple concatenation of the two inputs as done in ~\cite{sparse:to:dense:icra18} or adding a distance transformation map to indicate the location of the sparse values~\cite{depth:rgb:sparse:eccv18}.

We process both inputs individually with few convolutions before fusing them together. We find that this method works better than the ones used in~\cite{sparse:to:dense:icra18} and \cite{depth:rgb:sparse:eccv18}.
As to the convolutional network, we again use an encoder-decoder type of network architecture. The encoder takes the form:
\vspace{-2mm}
\begin{equation}
\label{eq:encoder:ID}
    X=f_{\text{encoder}}(I, \bar{D}),
\end{equation}
and the decoder is the same as defined in Eq.~\ref{eq:decoder}. As to the loss functions, in addition to the view-synthesis loss (Eq.~\ref{eq:vsloss:final}) and the edge-aware smoothing loss (Eq.~\ref{eq:smooth}) used for self-supervised depth prediction task, the berHu loss is also used but applied to the input sparse depth map $\bar{D}_t$ and its binary mask $\bar{M}(x,y)$.

The total loss for self-supervised depth completion is
\begin{equation}
\label{eq:total_loss}
    \mathcal{L}_{\text{self\_comp}} = \lambda\mathcal{L}_{berHu}^{\text{sparse}} + \mu\mathcal{L}_{vs} +  \nu\mathcal{L}_{smooth}.
\end{equation}
 Sparsity loss can act as supplemental loss to View-synthesis loss. The self-supervision from sparsity loss can handle scale ambiguities and textureless regions. It also stabilizes the training process and helps to converge faster. On the other hand, the view-synthesis loss is computed densely and can alleviate the effect of noise in the sparse depth maps.

We evaluate our method with three types of sparse patterns. Following the literature~\cite{sparse:to:dense:icra18}, our first pattern denoted by $\bar{D}^{\text{rand}}$ is created by randomly sampling pixels from ground truth depth image. The second pattern denoted by $\bar{D}^{\text{line}}$ is obtained by sampling the scanning lines ground truth depth images at a constant stride. The third pattern is the dense depth map $D$ itself, which is still sparse compared to images.

\subsection{Learning Time Series of Depth Maps}
\label{sec:monocular:video}
This section presents a framework to extend the three groups of methods such that they both train and test on time series of data. We formulate the depth recovery problem as a translation problem from a spatiotemporal sequence of multimodal data (i.e. images and sparse depth maps) to a spatiotemporal sequence of data (i.e. dense depth maps).
In order to model the spatiotemporal dependencies, we add the ConvLSTM module to the backbone network presented in the depth prediction section.
The ConvLSTM determines the future state of a certain cell in the grid from the inputs and past states of its local neighbors. As argued in~\cite{convlstm:nips:15}, if the hidden state is considered as the hidden representations of visual structures (objects), then ConvLSTM is able to capture motions of those visual components via its transitional kernels. Similarly, for the task of learning depth maps from monocular videos, we try to capitalize on temporal information to boost performance.
The correlation of the geometry of the scene and the perceived visual stimuli along motion trajectories should be captured and exploited.
Another well established approach to exploit temporal information is by concatenating multiple frames at the input. However, these approaches don't scale to longer sequences 
and require expensive 3D convolutions. We argue that long sequences are potentially beneficial for depth estimation from video combined with online learning methods.

Our encoder-decoder network, defined in Eq.~\ref{eq:encoder} and Eq.~\ref{eq:encoder:ID}, generates feature representations at varying levels. The output by the encoder $X$ is chosen as the input to our ConvLSTM. The choice is made due to the compactness of $X$ and its high information density, which leads to a more efficient optimization for the ConvLSTM.
More specifically, the learning process at frame $t$ starts with spatial convolutions with the encoder to get $X_t$, which is followed by temporal convolutions with the ConvLSTM
\vspace{-2mm}
\begin{equation}
\label{eq:concatenation}
    H_t,C_{t} = f_{\text{ConvLSTM}}((X_t \oplus H_{t-1}),C_{t-1}),
\end{equation}
and then followed by spatial convolutions with the decoder, such that $\hat{D} = f_{\text{decoder}}(H_t).$
The whole network is trained in an end-to-end manner. Its pipeline is sketched in Fig.~\ref{fig:pipeline} for the most complicated task self-supervised depth completion. The pipelines for supervised depth prediction and self-supervised depth prediction can be inferred  according to their inputs and loss functions.

\section{Training Framework}

\begin{figure}[!t]
\centering
\includegraphics[width=0.7\linewidth,angle=0]{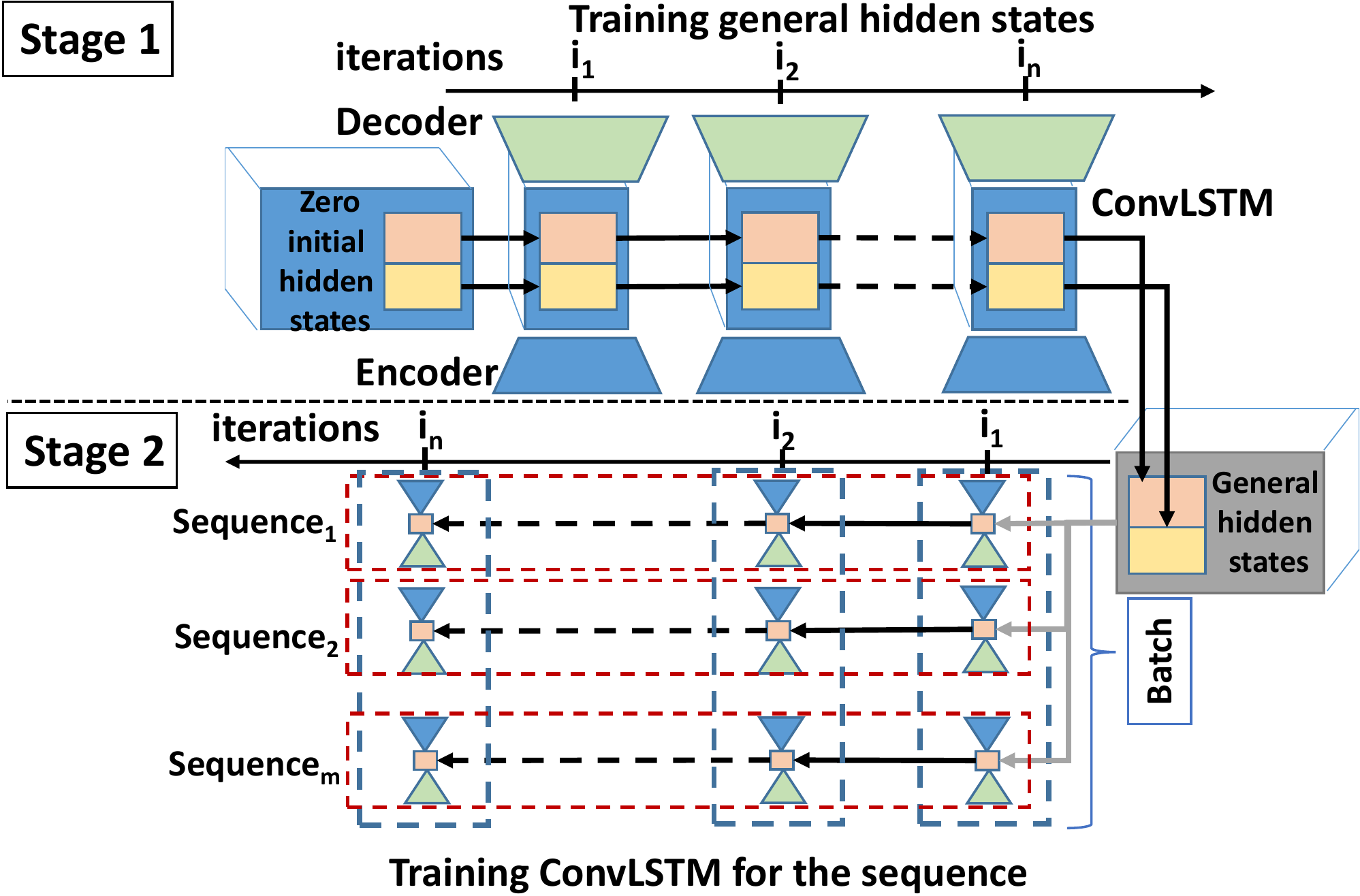}
\caption{Training procedure for hidden state of ConvLSTM.}
\label{fig:trainingLSTM}
\vspace{-3mm}
\end{figure}

\begin{figure*}[!htb]
\centering
\includegraphics[width=1.0\textwidth, trim=0 34 0 0, clip]{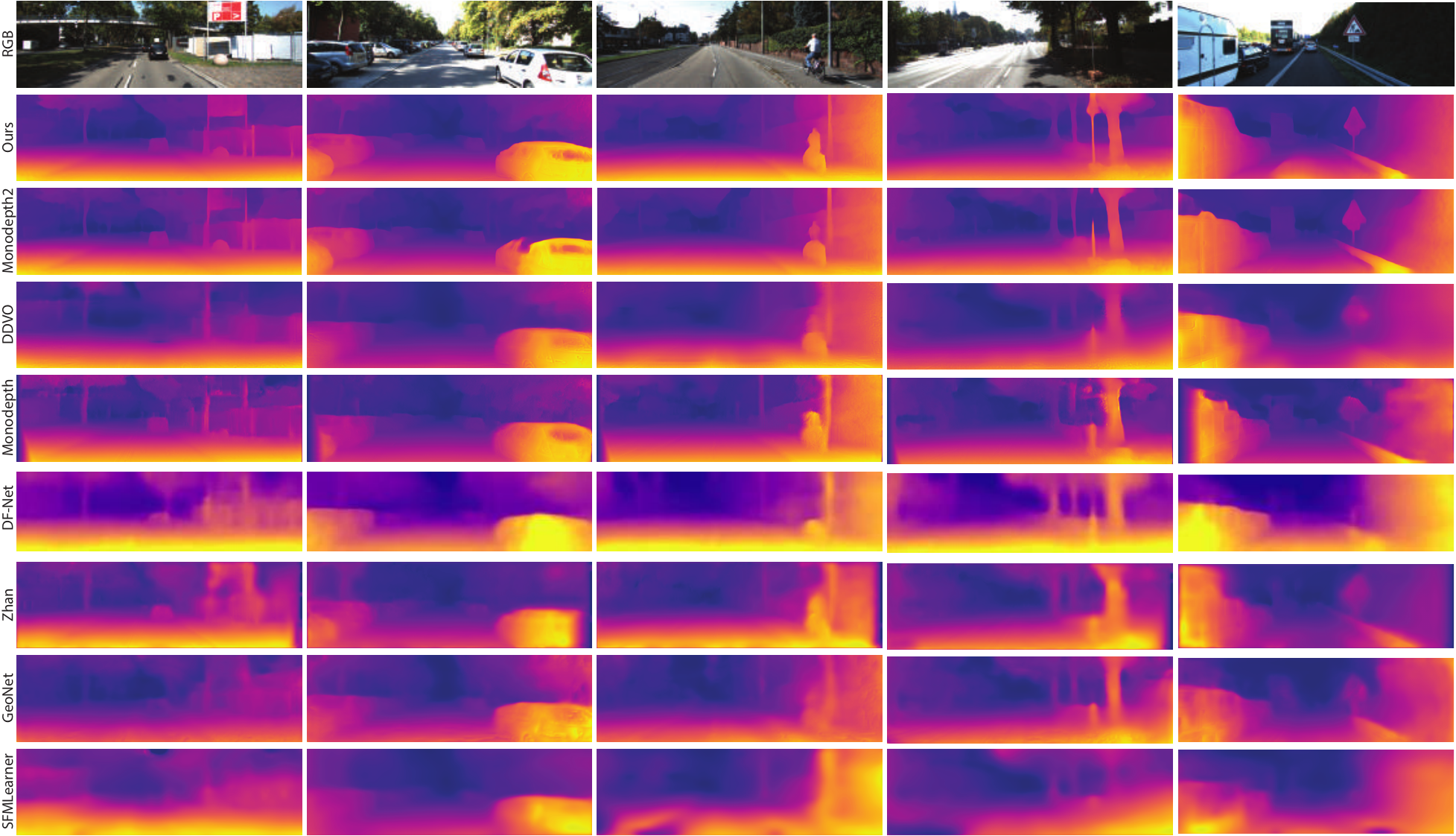}
    \caption{\textbf{Qualitative results on the KITTI Eigen split.} Our self-supervised recurrent method generates more accurate depth maps compared to  competing methods~\cite{monodepth2:iccv19, ddvo:cvpr18, unsupervised:depth:left:right:cvpr17, dfnet:eccv18, depth:odometry:cvpr18, geonet:cvpr18}, especially for small objects like poles, tree trunks and traffic signs. }
\label{fig:visual:results}
\vspace{-2mm}
\end{figure*}

\subsection{Network Architecture}
\label{sec:network:architecture}
The network architecture consists of a depth prediction network and a pose network. The encoder branch of both networks contains separate ResNet-18 modules~\cite{resnet:cvpr16}. The decoder unit of the depth network contains four upconvolutional blocks inspired from DispNet~\cite{dispnet:cvpr16}. The output of the encoder is connected with the ConvLSTM~\cite{convlstm:nips:15} module. The ConvLSTM module receives hidden state $H_{t-1}$ and cell state $C_{t-1}$ of ConvLSTM from the previous frame $t-1$ (details in Sec.~\ref{sec:monocular:video}). The output of this network is disparity extracted at different spatial resolution from each unit of decoder. The pose decoder consist of stack of Conv(1) and Conv$(3)$ blocks and produces a 6 element vector representing axis angle and translation. For simplicity, we combine the hidden representation $H$ and cell state $C$ in the ConvLSTM and refer to it further as the~\textit{hidden state}. The~\textit{initial hidden state} refers to the initialization of the \textit{hidden state}.

\subsection{ConvLSTM Training Strategy}
\label{sec:training:lstm}
In vanilla LSTM based network training, the default strategy is to initialize the hidden state to zero. This is a well established practise in sequence to sequence learning models. Here, the impact of the initial hidden state is either trivial or the length of the sequence is relatively long compared to the size of the hidden state. In case of training a ConvLSTM based network with monocular video, the size of the feature maps in the bottleneck increases drastically. This demands increased capacity of the hidden state due to the concatenation (Eq.~\ref{eq:concatenation}).
With more learnable parameters and shorter sequence lengths during training, the effect of the initial hidden state becomes dominant. In our training procedure, we try to mitigate the effects of a zero initial hidden state.
The training is divided into two stages as depicted in Fig.~\ref{fig:trainingLSTM}. In the first stage, the initial hidden state is considered as a learnable parameter. The training begins with the hidden state initialized with zeros. Further on, for every iteration we first load the initial hidden state and backpropagate through the hidden state to update the initial hidden state of the ConvLSTM. The training is performed on fully randomized batches. This procedure allows us to learn a general initial hidden state.
In the second stage, the trained initial hidden state is used at the start of every new sequence. Hence, the training is performed on the video sequences as opposed to the previous stage. This means that the weights of ConvLSTM module are adjusted based on the pretrained hidden state from the first stage. The training weights from the second stage along with learned initial hidden state from the first stage are used during testing. It enables the ConvLSTM to adapt to the sequence faster, resulting in superior performance as shown in Sec.~\ref{subsec:hidden_state}. The weights are updated for every frame to optimize for training speed and memory footprint (i.e. truncated BPTT~\cite{tbtt:90} with window size of 1).
Although this technique is effective for a large dataset with long sequences, it can result in overfitting when updating the weights per single image.
To alleviate this issue, we also train in batches during the second stage. This allows us to update the batch statistics and average the gradients. This greatly benefits the generalization. Not all sequences have the same length, therefore we first divide the original video sequences into smaller ``sub-sequences" for training, and load multiple random ``sub-sequences" in parallel afterwards. This whole procedure is depicted in Fig.~\ref{fig:trainingLSTM}. The batch is shown by the dotted blue rectangle. The influence of the sequence length is evaluated in Sec.~\ref{subsec:hidden_state}.

\begin{table*}[!htb]
\vspace{-1mm}
\centering
\caption{Results for self-supervised depth prediction.}
\setlength{\tabcolsep}{2pt}
\begin{tabular}{@{\extracolsep{2pt}}l c c c c c c c}
\toprule[1pt]
\textbf{Method} & $\downarrow$ \textbf{RMSE} & $\downarrow$ \textbf{RMSE(log)} & $\downarrow$ \textbf{Abs Rel Diff} & $\downarrow$ \textbf{Sq Rel Diff}  & $\uparrow$ $\boldsymbol{\delta<1.25}$  & $\uparrow$ $\boldsymbol{\delta<1.25^{2}}$ & $\uparrow$ $\boldsymbol{\delta<1.25^{3}}$ \\
 \midrule[0.3pt]
SFMLearner \cite{unsupervised:depth:ego:cvpr17} & 6.709  &  0.270 & 0.183 & 1.595 & 0.734 & 0.902 & 0.959 \\
DDVO~\cite{ddvo:cvpr18} & 5.583 & 0.228 & 0.151 & 1.257 &  0.810 & 0.936 & 0.974\\
GeoNet~\cite{geonet:cvpr18} & 5.567 & 0.226 & 0.149 & 1.060 &  0.796 & 0.935 & 0.975\\
CC~\cite{ranjan:cvpr19} & 5.464 & 0.226 & 0.148 &  1.149 &  0.815 &  0.935 &  0.973 \\
EPC++~\cite{epc++:18} & 5.350 & 0.216 & 0.141 & 1.029 & 0.816 & 0.941 & 0.976 \\
Struct2depth (w/o ref.)~\cite{struct2depth:aaai19} & 5.291 & 0.215 & 0.141 & 1.026 & 0.816 & 0.945 & 0.979 \\
GL-Net (w/o ref.)~\cite{glnet:chen:cvpr2019} & 5.230 & 0.210 & 0.135 & 1.070 & 0.841 & 0.948 & 0.980 \\
Monodepth2~\cite{monodepth2:iccv19} & 4.863 & 0.193 & 0.115 & 0.903 & 0.877 & 0.959 & 0.981 \\
\textbf{Ours} (Average over 5 runs) & \textbf{4.730} & \textbf{0.188} & \textbf{0.112} & \textbf{0.863} & \textbf{0.879} & \textbf{0.960} & \textbf{0.981} \\
\textbf{Ours} (Best) & \textbf{4.650} & \textbf{0.187} & \textbf{0.111} & \textbf{0.821}  & \textbf{0.883} & \textbf{0.961} &	\textbf{0.982} \\
\bottomrule[1pt]
\end{tabular}
\label{tab:eop-expt}
\vspace*{-5mm}
\end{table*}

\subsection{Implementation Details}

\noindent\textbf{Training Details}. The weights $\nu$ and $\lambda$ are respectively used for the smoothing and sparsity loss in Eq.~\ref{eq:total_loss}. The former is set to $0.001$ while the latter is iteration dependent. In fact, $\lambda(i)$ changes during training to prevent overfitting on a low number of LiDAR points. We start with the view-synthesis loss and smoothing loss first and gradually increase the influence of the sparsity loss afterwards. This happens linearly with the number of iterations $i$, such that $\lambda(i) = 10^{-2}\cdot\min(1, 10^{-3}\cdot i)$. After all, the network should be prevented from learning the identity function in order to discover semantics and depth. For all experiments, we use a batch size of 12, with the Adam optimizer and a learning rate of $10^{-4}$. The images are resized to a resolution of 192x640 as in Monodepth2~\cite{monodepth2:iccv19} baseline.
Training takes 10 epochs for the first stage, while we finetune on sequences during the second stage for 20 epochs.  For the encoder, we used pretrained ImageNet~\cite{imagenet:nips12} weights. This is important to achieve competitive results as in \cite{epc++:18, yang:eccv18,monodepth2:iccv19, ranjan:cvpr19}. All the other weights are initialized with He initialization~\cite{he:iccv15}, except for the biases of the convolution layer at the forget gate. To make the ConvLSTM focus on the hidden state at the start of training, we set the biases to 1 as in~\cite{lstm:icml}. We replaced the Tanh activations in the ConvLSTM with ELU~\cite{elu} in order to match the scale with the output of encoder (Eq.~\ref{eq:concatenation}).

\section{Experiments}
To show the effectiveness of our approach, we address the previously defined conditions in Sec.~\ref{sec:approach}: 1) the supervised depth prediction setup with raw LiDAR ground truth, 2) the self-supervised depth prediction setup and 3) the self-supervised depth completion setup. For each case, numbers are reported on the KITTI dataset in order to evaluate with other monocular depth estimation methods. We consider Monodepth2~\cite{monodepth2:iccv19} as our baseline in the following self-supervised depth estimation experiments.

The supervised depth prediction and self-supervised depth completion experiments are evaluated on the Eigen split defined by Eigen \textit{et\ al.}~\cite{Eigen:2015}. This split contains 28 raw KITTI sequences for training, 5 sequences for validation and 28 sequences for testing, all with variable length. Our approach is not limited to the fixed sequence length adopted during training. To show this generalization towards longer sequences, we always evaluate on the complete video sequences during testing. Only for the self-supervised depth prediction task, we use a filtered version in order to leave out static frames, as defined by Zhou~\textit{et\ al.} \cite{unsupervised:depth:ego:cvpr17}. Furthermore, to achieve absolute depth, our predictions are rescaled with the median ground truth depth per frame, as done in previous works \cite{monodepth2:iccv19, struct2depth:aaai19, unsupervised:depth:ego:cvpr17, ranjan:cvpr19, epc++:18, ddvo:cvpr18}. It is worth noting that the predictions of the self-supervised depth completion setup do not require re-scaling since the sparsity loss (Eq.~\ref{eq:total_loss}) enables the predictions to be scale-aware. The quantitative results are reported on the selected 697 frames from the 28 test sequences unless mentioned otherwise. In all our experiments, we cap the maximum predictions of all networks to 80m.

\begin{table}[!bt]
    \centering
    \caption{Results of self-supervised depth completion.}
    \setlength{\tabcolsep}{3pt}
    \small{
    \begin{tabular}{@{\extracolsep{4pt}}l c c c}
        \toprule[1pt]  
     \textbf{Input} & \textbf{Method}  & $\downarrow$ \textbf{RMSE} & $\uparrow$ $\boldsymbol{\delta < 1.25}$  \\
         \midrule[0.3pt]  
     \multirow{2}{*}{$\bar{D}^{\text{rand}}_{500}$}  
      & Image-based & 2.885 & 0.970 \\
      & Recurrent & \textbf{2.738} & \textbf{0.973} \\
    \hline
      \multirow{2}{*}{$\bar{D}^{\text{line}}_{8}$}
      &  Image-based &  2.750 &  0.962\\
      & Recurrent & \textbf{2.586}  & \textbf{0.968}  \\
          \hline
     \multirow{3}{*}{$\bar{D}^{\text{line}}_{64}$}
    & Ma \textit{et\ al.} \cite{ma:self:sparse:icra19} & 1.922 & 0.985 \\
    & Image-based & 1.653 & 0.988 \\
      & Recurrent & \textbf{1.592} & \textbf{0.990 }\\
        \bottomrule[1pt]
    \end{tabular}}
    \label{tab:diffmod-expt}
    \vspace{-5mm}
\end{table}

\subsection{Analysis}
In this section, we discuss the qualitative and quantitative results. We achieve a new state-of-the art for both self-supervised setups, proving its effectiveness.

\noindent \textbf{Self-Supervised Depth Prediction} The results in Table~\ref{tab:eop-expt} shows that our method outperforms recent state-of-the-art methods. We improve over our baseline by a relatively large margin (-0.133m RMSE). The qualitative results are depicted in Fig.~\ref{fig:visual:results}. The baseline method only use short-range video information when computing the view-synthesis loss, while our method leverages longer-range temporal information. Highly reflective surfaces (e.g. mirrors) or dynamic objects can still cause problems due to limitations of the self-supervision loss. We do not report results obtained by refining the model during test time using test images as in \cite{struct2depth:aaai19} \cite{glnet:chen:cvpr2019}.

\begin{table}[!b]
\vspace{-5mm}
    \centering
    \caption{Results of supervised depth prediction.}
    \setlength{\tabcolsep}{2pt}
    \small{
    \begin{tabular}{@{\extracolsep{4pt}}c c c c}
    \toprule[1pt]  
     \textbf{Method} & $\downarrow$ \textbf{RMSE} & $\downarrow$ \textbf{Abs Rel Diff} & $\uparrow$ $\boldsymbol{\delta < 1.25}$  \\
\midrule[0.3pt]
Eigen \textit{et\ al.}~\cite{Eigen:2014} fine & 7.156 & 0.215 & 0.692 \\
Liu \textit{et\ al.}~\cite{depthfield:single:image:pami16} & 6.986 & 0.217 & 0.647 \\
Kumar \textit{et\ al.}~\cite{lstm:cvprw18} & 5.187 & 0.137 & 0.809 \\
Wang \textit{et\ al.}~\cite{wang-lstm:cvpr19}* & 5.106 & 0.128 & 0.836 \\
Kuznietsov \textit{et\ al.}~\cite{semi:depth:cvpr17} & 4.621 & 0.113 &  0.862\\
Yang \textit{et\ al.}~\cite{yang:eccv18} & 4.442 & 0.097 & 0.888 \\
Guo \textit{et\ al.}~\cite{guo:eccv18} & 4.422 & 0.105 & 0.874 \\
Zhang \textit{et\ al.}~\cite{zhang:iccv19:exploiting} & 4.139 & 0.104 & 0.883 \\
Fu \textit{et\ al.}~\cite{dorn:cvpr18}* & \textbf{\underline{3.714}} & \textbf{\underline{0.099}} & \textbf{\underline{0.897}} \\
\textbf{Ours (w/o recurrent)} & 4.320 & 0.113 & 0.874 \\
\textbf{Ours} & \textbf{4.148} & \textbf{0.102} & \textbf{0.884}\\
    \bottomrule[1pt]
    \end{tabular}}
    \label{tab:sup-expt}
\end{table}

\begin{table}[!ht]
\caption{Results on Eigen split with LiDAR supervision evaluated for different activation layers and sequence lengths.}
\centering
\setlength{\tabcolsep}{-0.2pt}
\small{
\begin{tabular}{@{\extracolsep{3pt}} @{\extracolsep{4pt}}c c c c c c c c}
\toprule[1pt]
\textbf{Activat.} & \textbf{Frames}  & $\downarrow$ \textbf{RMSE} &  $\downarrow$ \textbf{Abs Rel}  & $\uparrow \boldsymbol{\delta<1.25}$  & $\uparrow$ \textbf{$\boldsymbol{\delta<1.25^2}$}  \\
 \midrule[0.3pt]
None & 30 & 4.210 &	0.108 &	0.881 &	0.965 \\
Tanh & 30 & 4.370 &	0.115 &	0.874 &	0.964 \\
ReLU & 30 & 4.173 &	0.104 &	\textbf{0.884} &	0.965 \\
ELU  & 30 &  \textbf{4.148} & \textbf{0.102} & \textbf{0.884} &	\textbf{0.966} \\
\hline
ELU  & 15  & 4.234 & 0.107 & 0.882 & 0.964 \\
ELU  & 50  & 4.155 & 0.103 & \textbf{0.884} & \textbf{0.966} \\
ELU  & 100 & 4.170 & 0.103 & 0.883 & 0.965 \\
\bottomrule[1pt]
\end{tabular}}
\label{tab:activations}
\end{table}

\noindent\textbf{Self-Supervised Depth Completion}. We perform experiments with three sparse patterns as defined in Sec.~\ref{sec:self:depth:comp}. For all patterns, we compare our recurrent method to its image-based counterpart. Since our method also uses the input patterns of $64$ LiDAR lines by Velodyne HDL-64E, we report the results on the common $652$ images of Eigen set~\cite{Eigen:2015} and the KITTI depth benchmark dataset for which the corrected ground truth are available. The results are reported in Table~\ref{tab:diffmod-expt}. The  table shows that  the  proposed  recurrent framework outperforms its image-based  counterpart for all three different sparse patterns. Our method also outperforms the state-of-the-art self-supervised depth completion method~\cite{ma:self:sparse:icra19}. In addition to the use of longer-range temporal information, the better performance can also be attributed to the good features of our baseline Monodepth2: pixel-wise masking of the loss and  better strategy to handle occlusions.

\noindent \textbf{Supervised Depth Prediction}. The quantitative results are shown in the Table~\ref{tab:sup-expt}. Performing regression towards the re-projected LiDAR points is not ideal, due to the noisy LiDAR data~\cite{van:sparse:mva19}. We hypothesize that our recurrent approach can add consistency and produce more accurate predictions (-0.172m RMSE). This can be supported by 1) only marginal improvement is observed, when training on the corrected KITTI ground truth (dense), 2) $\delta_1$ is considerably higher (+1\%)  than in our baseline supporting our claim. In this setup we are still outperformed by~\cite{dorn:cvpr18}. However, they use a complex network architecture (ResNet-101) with fully connected layers with inference time of 500 ms. Thi is not applicable to real-time depth prediction tasks, as in autonomous driving compared to our inference time of 10 ms. Note that, here we re-evaluate~\cite{dorn:cvpr18} with our setting. Also, we report corrected result of~\cite{wang-lstm:cvpr19} in supervised setting based on predictions provided by the authors and omit erroneous results for the unsupervised case in~\cite{wang-lstm:cvpr19}.
\begin{figure}[!ht]
\centering
\includegraphics[width=0.95\linewidth, angle=0]{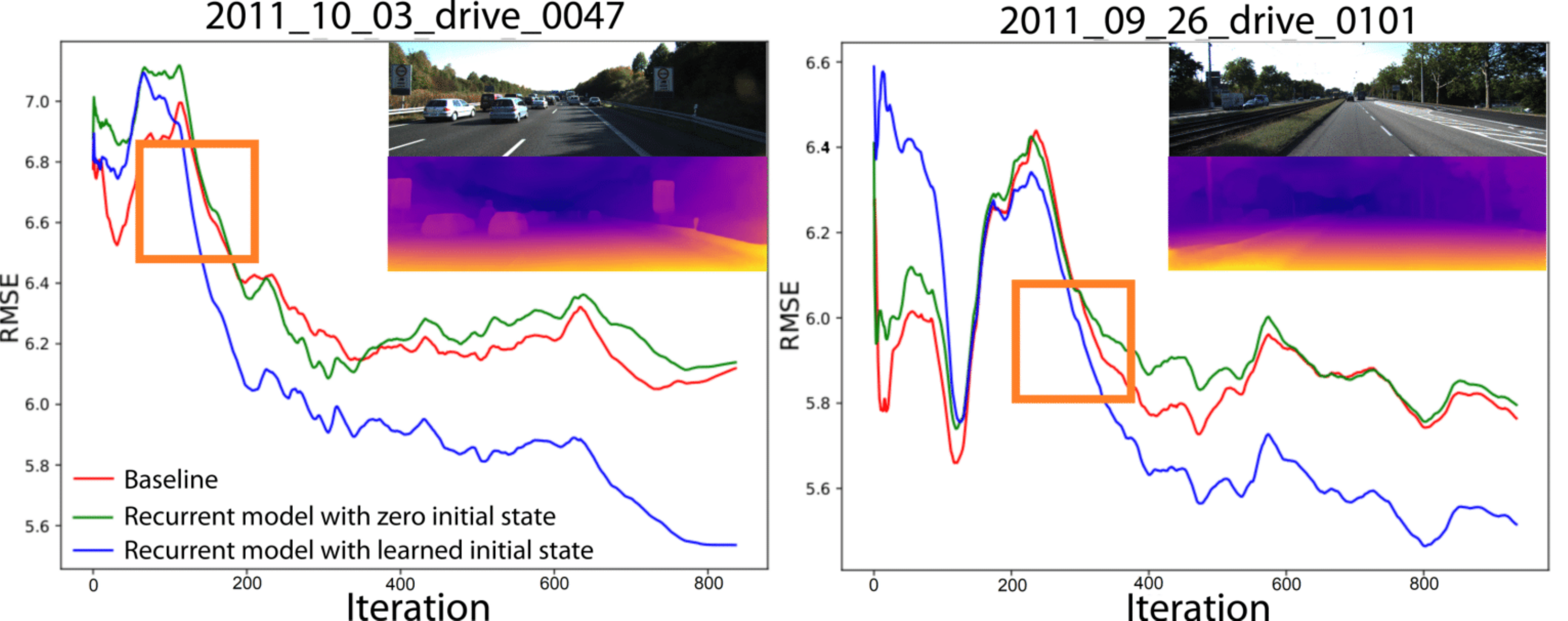}
\caption{Accumulated average RMSE (RMSE averaged over all previous frames) for three KITTI video sequences.}
\label{fig:online:plots}
\end{figure}

\begin{figure}[tb]
\vspace{-4mm}
\centering
\includegraphics[width=0.7\linewidth]{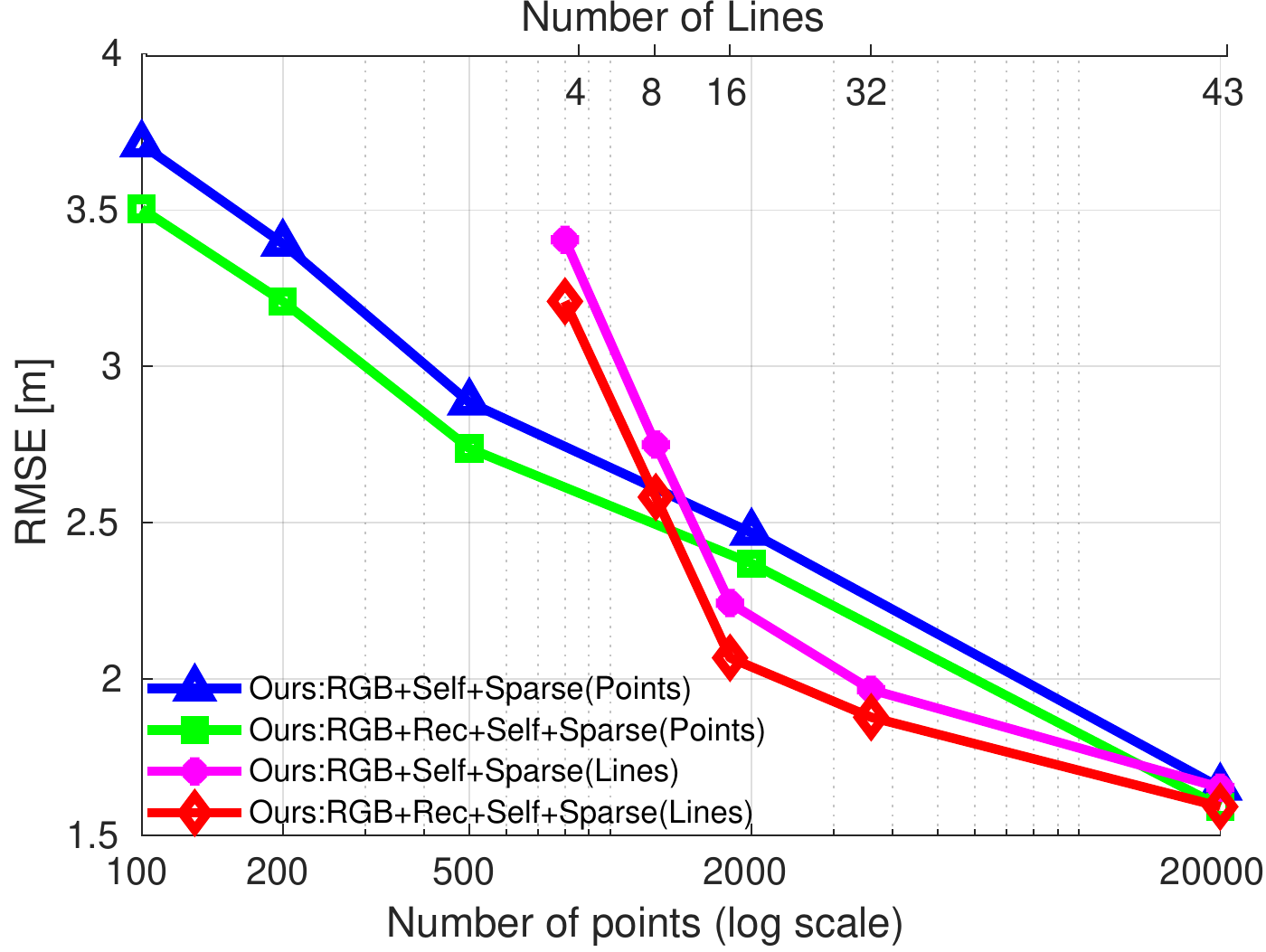}
\caption{Performance of our method as a function of the number of sparse points and the number of scanning lines.}
\label{fig:sparsity:plot}
\end{figure}

In Fig.~\ref{fig:sparsity:plot}, we evaluate the performance of our method as a function of  the number of sparse points in $\bar{D}^{\text{rand}}$ and the number of scanning lines in $\bar{D}^{\text{line}}$. As expected, our method benefits from having denser depth samples as the inputs for both evaluated scenarios. Our recurrent framework is able to exploit the spatial and temporal structures of the scenes in the case of sparse points and scanning lines as well and consistently outperforms its frame-based counterpart. The improvement in RMSE score drops as number of input points increases. When supervision from LiDAR gets stronger, the reliance on other sources decreases.

\subsection{Ablation Study}
\label{subsec:hidden_state}
\noindent\textbf{Pretrained Initial Hidden State}. We compare zero-initialized training strategy with ours and report the results over KITTI sequences in Fig.~\ref{fig:online:plots}. The figure shows that training the initial hidden state as a variable is more effective than using zeros as initial states. The pretrained initial state helps to speed up adaptation and improves generalization at the beginning of a sequence. For example, we observe better initial predictions for the recurrent model with a learnable initial hidden state in (eg. sequence $47$) Fig.~\ref{fig:online:plots} than with the zero initialization. However, in sequence $101$ we show a counter example. Interestingly, the network is still able to outperform the baseline over time in this sequence. As one can see from the figure, our training strategy achieves considerable and consistent improvement over the zero-initialized training after certain number of frames. The improvements are noticeable when the car is stationary or in constant motion. The re-scaling factor in the self-supervised setup varies less in those regions.


\begin{figure}[tb]
\centering
\includegraphics[width=\linewidth, angle=0]{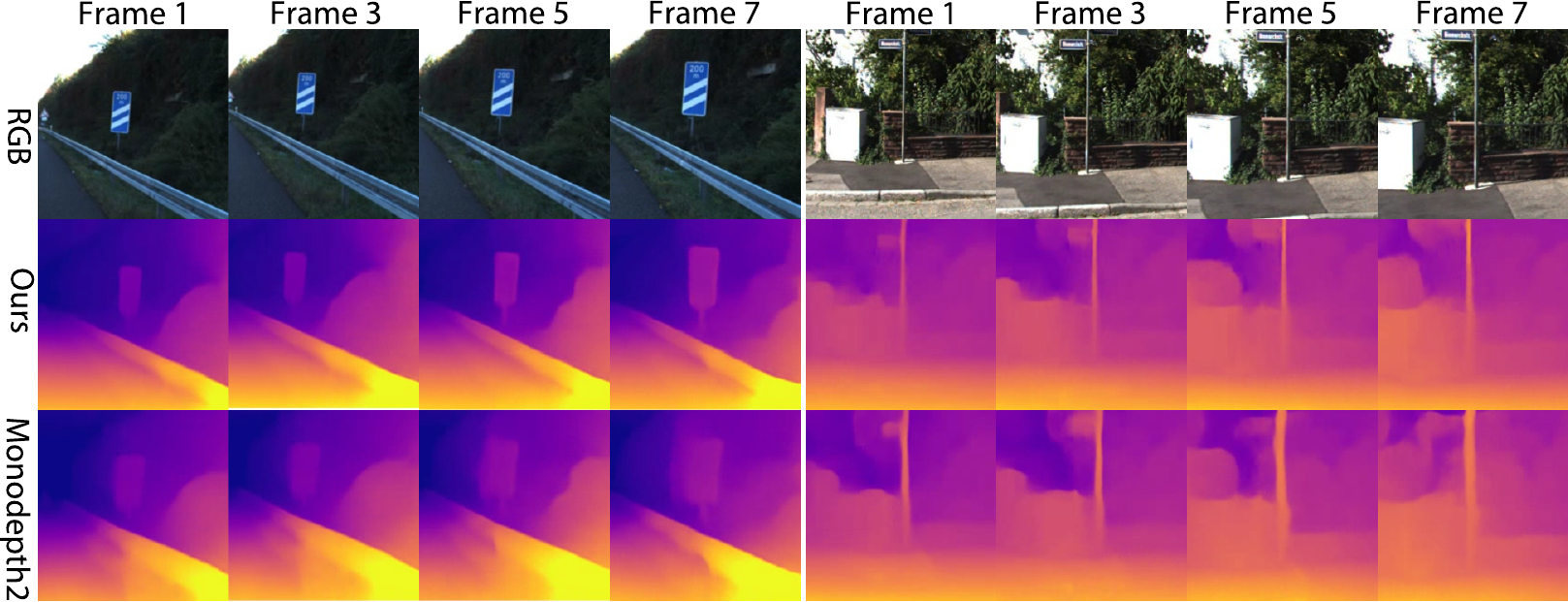}
\caption{Recurrent method demonstrates better temporal Consistency on KITTI video sequences over image based method.}
\label{fig:temporal:plots}
\end{figure}

\begin{table}[h]
\vspace{-2mm}
    \centering
    \caption{Rel. improvement of Recurrent method compared to baseline}
    \setlength{\tabcolsep}{3pt}
    \small{
    \begin{tabular}{c c c c}
        \toprule[1pt]  
    \textbf{Method}  & $\Delta$ \textbf{RMSE} & $\Delta$ $\boldsymbol{\delta < 1.25}$ & $\Delta$ \textbf{ARTE}  \\
         \midrule[0.3pt]  
    Self sup. depth prediction & -0.133 & 0.002 & -0.010\\
    Self sup. depth completion & -0.061 & 0.002 & -0.005\\
        \bottomrule[1pt]
    \end{tabular}}
    \label{tab:delta}
    \vspace{-1mm}
\end{table}

\noindent\textbf{Temporal consistency}. We also evaluate our method for temporal consistency Fig.~\ref{fig:temporal:plots}. The quantitative metrices defined by \cite{zhang:iccv19:exploiting} are not suitable for Datasets with sparse ground truth. We define our evaluation metric, Absolute Relative Temporal Error(ARTE), as follows: $ \frac{1}{T} \sum\nolimits_{i \in T} \frac{|(| \hat{D}_{i} - \hat{D}_{i-1} | - | D_{i} - D_{i-1} | )|}{| D_{i} - D_{i-1} | + \epsilon} $.
We set $\epsilon$ to 0.001 and evaluate our predictions for self-supervised depth estimation first. Compared to the image-based baseline, our frame-recurrent method reduces the ARTE from 0.1401 to 0.1297. This improvement is also reflected in the RMSE~(Table~\ref{tab:delta}). Additionally, we perform the same experiment for depth completion. The improvement over the image-based baseline  is lower, i.e. 0.005. 
This is in line with the numbers in Table~\ref{tab:diffmod-expt}, indicating that temporal consistency is more beneficial when fewer LiDAR points are available as input.



\noindent\textbf{Activation Functions}.
The Tanh activation was introduced in LSTMs to deal with vanishing gradient problem in very long sequences. Intuitively, we expect to see better results with activations which preserve the input range. This is not the case for Tanh. The scale of Tanh does not match the scale of the input, a necessity for concatenation of the input with the hidden state. We evaluate the effect of different activation functions operating on the states inside the ConvLSTM. The results are shown in the Table~\ref{tab:activations}.  In our case, Tanh is inferior to ReLU, ELU~\cite{elu} and even to no activation. The lowest RMSE score is achieved with ELU.

\noindent\textbf{Training Sequence Length}. The effect the training sequence length is shown in Table~\ref{tab:activations}. Experiments show that a sequence length of $30$ strikes a good balance between performance and cost. Using short sequences leads to worse results; using longer ones does not boost the performance further. Training on very long sequences, 100 or higher, achieves similar results. This means that the ConvLSTM is able to capture temporal information by propagating the hidden state through the whole sequence.

\section{Conclusion}
This work has introduced a novel method for estimating time series of depth maps with monocular video and optionally sparse depth.  Our method exploits the spatio-temporal structures over data frames both at train and test time for accurate depth maps. Specifically, a recurrent framework has been developed and evaluated for three different tasks: supervised depth prediction, self-supervised (SS) depth prediction and self-supervised depth completion. For both SS scenarios, we outperform current SOTA methods significantly.


\bibliographystyle{./IEEEtran}
\bibliography{./IEEEabrv,./IEEEexample}

\end{document}